\documentclass[preprint,5p]{elsarticle}

\usepackage{booktabs}
\usepackage{amssymb}
\usepackage{amsmath, amsfonts}
\usepackage{float}
\usepackage{stfloats}
\usepackage{tabularx}
\usepackage{placeins}

\usepackage{caption}
\usepackage{graphicx}
\usepackage{booktabs}
\usepackage[table]{xcolor}
\usepackage{makecell}
\usepackage{array}
\usepackage[T1]{fontenc} % T1 font encoding kullanır, daha kaliteli çıktı
\usepackage{lmodern}     % Latin Modern yazı tiplerini kullanır

\usepackage[colorlinks=true, linkcolor=blue, citecolor=blue, urlcolor=blue]{hyperref}

\usepackage[nameinlink,capitalise]{cleveref}

\let\cite\citep

\journal{Knowledge-Based Systems}

\begin{document}

\begin{frontmatter}

\title{Predictive Modeling and Anomaly Detection in Large-Scale Web Portals Through the CAWAL Framework}
%%\tnotetext[label1]{}

\author[a,b]{Özkan Canay\corref{cor_a}}

\affiliation[a]{organization={Sakarya University of Applied Sciences},
	addressline={Vocational School of Sakarya, Dept. of Computer Tech.},
	postcode={54290},
	city={Sakarya},
	country={Türkiye}}
\affiliation[b]{organization={Sakarya University},
	addressline={Institute of Natural Sciences, Dept. of Computer and IT Engineering},
	postcode={54050},
	city={Sakarya},
	country={Türkiye}}
\ead{canay@subu.edu.tr}
\ead[url]{https://canay.subu.edu.tr/}
\cortext[cor_a]{Corresponding author}


\author[c,d]{Ümit Kocabıçak}
\affiliation[c]{organization={Turkish Higher Education Quality Council},
	postcode={06800},
	city={Ankara},
	country={Türkiye}}
\affiliation[d]{organization={Sakarya University},
	addressline={Faculty of Computer and IT Engineering, Dept. of Computer Eng.},
	postcode={54050},
	city={Sakarya},
	country={Türkiye}}

\begin{abstract}
The Web Usage Mining (WUM) process relies heavily on web server logs, which are limited in data diversity and quality. As user interactions become more complex within large-scale web server architectures and multi-service web portals, these limitations present significant challenges for conventional WUM and Web Analytics. To address these challenges, CAWAL was developed as an innovative model that integrates application logs with web analytics to collect comprehensive and detailed interaction data. This study explores the use of enriched session and pageview data obtained through the CAWAL framework to enhance WUM processes, improve the accuracy of prediction models, and optimize anomaly detection. Analyses show that advanced machine learning models, such as Gradient Boosting and Random Forest, applied to these enriched datasets, achieved over 92\% accuracy in predicting user behavior and significantly improved anomaly detection. The framework that provides advanced data integration also accelerates the mining process by eliminating the preprocessing stage. Research findings demonstrate that the framework offers a scalable and efficient solution, providing valuable insights into user interactions and system performance, making it highly suitable for optimizing modern web portals.
\end{abstract}

\begin{graphicalabstract}
\includegraphics[scale=1.35]{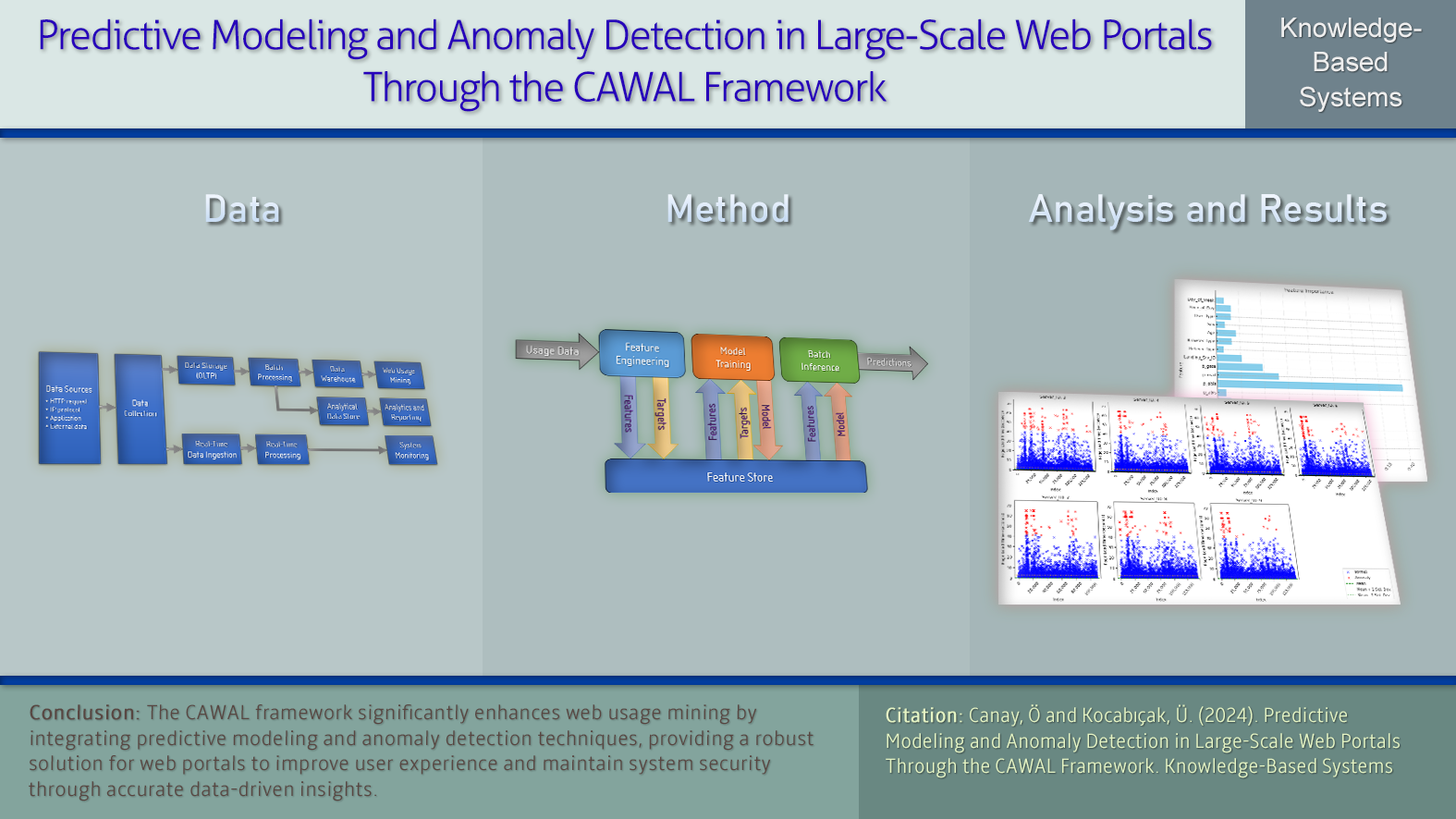}
\end{graphicalabstract}

\begin{highlights}
\item CAWAL eliminates the preprocessing phase, speeding up WUM analysis.
\item Enriched datasets enable ML models to achieve over 92\% prediction accuracy.
\item The model enhances anomaly detection in multi-server web farm architectures.
\item The method provides a comprehensive data foundation for portal optimization.
\end{highlights}


\end{frontmatter}

%\linenumbers
\sloppy

\section{Introduction}\label{sec:1}

Web Usage Mining (WUM) is the process of analyzing user interactions on websites to extract meaningful and valuable insights from their behavior. Web logs play a critical role by providing essential data such as navigation paths, pages visited, and interaction durations, enabling the analysis of user behaviors on websites \cite{Hopipah2021}. Accurate analysis of user interactions is crucial for strategic decision-making, especially in fields like e-commerce, online education, and security \cite{Kumar2022}. However, the growing volume of data and the increasing complexity of user interactions challenge the capacity of traditional methods to process large datasets efficiently. Hence, integrating machine learning and data mining techniques into WUM offers significant opportunities, particularly in predictive modeling and anomaly detection, while also introducing new challenges \cite{Dubey2024}.

Data preprocessing, one of the most critical stages in WUM, involves cleaning and organizing weblogs to extract meaningful information. However, traditional data preprocessing methods are time-consuming and complex, especially for large datasets \cite{Alshdaifat2021}. For example, the use of social network analysis and frequent pattern mining to discover valuable information from extensive web data was proposed \cite{Leung2021}, while fuzzy techniques and clustering were focused on to understand user behavior in large datasets \cite{Gangadwala2022, Zain2017}. Despite these advancements, the need for more comprehensive and automated data processing techniques is growing.

Predictive modeling, a widely used WUM application, is a crucial method for forecasting future user activities on websites. In recent years, a study successfully applied Long Short-Term Memory (LSTM) networks to predict e-commerce users' shopping intentions with high accuracy \cite{Diamantaras2021}. Similarly, another recent study achieved high success in web page prediction using a chicken swarm optimization model based on neural networks \cite{Gangurde2022}. These models contribute to strategic decision-making by predicting users' shopping tendencies and browsing habits. However, the accuracy of such predictive models is directly linked to the scope and richness of the datasets used. Due to their limited data coverage, web server logs often constrain these models' performance.

Another important application of WUM is anomaly detection, which focuses on identifying abnormal user behaviors and is commonly employed to detect security threats and performance issues. For example, a multi-behavior fusion-based security system has been shown to achieve high accuracy in detecting anomalies and misuse within computer networks \cite{Gupta2021}. In a similar vein, an automated analysis method utilizing NGINX logs has been developed to detect user anomalies in large web server logs, providing a valuable approach for enhancing cybersecurity in network analysis \cite{Benova2023}.

This study investigates using datasets generated by the CAWAL framework \cite{Canay2024} to improve the performance of predictive modeling and anomaly detection techniques in web usage mining. CAWAL integrates web analytics data with application logs, providing more extensive and enriched data sources. In scenarios where web server logs are insufficient, combining detailed data such as session information, page view records, user profiles, and interaction history is expected to enhance the accuracy of predictive models and improve the efficiency of anomaly detection processes.

The hypothesis (H1) that the enriched datasets provided by the CAWAL framework will enhance the accuracy of machine learning-based prediction models in large-scale, multi-server architectures is tested in this study. CAWAL is expected to improve the performance of WUM processes by delivering more reliable results for predictive models. Additionally, the hypothesis (H2) that the framework will optimize anomaly detection in multi-server systems, thereby improving system efficiency and security, is also tested. These hypotheses are explored within this study's scope to evaluate the framework's performance in WUM processes.

The main contributions of this study are as follows:

\begin{enumerate}
	\item  The acceleration of WUM processes by eliminating the preprocessing step using CAWAL-provided data.
	
	\item  Improvement in the accuracy of predictive models and their application to large datasets.
	
	\item  Optimization of anomaly detection processes in multi-server and multi-domain architectures, such as web farms.
	
	\item  Provision of a more comprehensive data infrastructure for optimization and decision-making processes in web portals.
\end{enumerate}

The remainder of this paper is structured as follows: \cref{sec:2} reviews existing approaches in web usage mining and discusses the innovations introduced by the CAWAL framework. \cref{sec:3} details the framework's architecture, data flow and processing steps, and preparations for machine learning. \cref{sec:4} analyzes the prediction models and anomaly detection using enriched data and examines the experimental results. \cref{sec:5} provides a thorough discussion of the findings, and finally, \cref{sec:6} offers a general evaluation of the study, conclusions, and suggestions for future research.

\section{Related work}\label{sec:2}

Web Usage Mining is the process of analyzing weblogs to extract meaningful patterns from users' online interactions. The data, such as users' browsing habits, pages visited, and session durations, form the core of WUM's information sources \cite{Yau2020}. This process is supported by machine learning and data mining techniques to handle and analyze large datasets. \cref{tab1} summarizes recent studies on web usage mining, focusing on prediction and anomaly detection, along with the methods and techniques used.

\begingroup
\setlength{\extrarowheight}{3pt}  % Table row spaces
\begin{table*}[htb]
	\setlength{\parindent}{0pt} % Tablo girintisini sıfırla
	\centering
	\caption{Recent studies on web usage mining, prediction, and anomaly detection methods.}
	\label{tab1}
	\small
	\begin{tabular}{ m{0.465\textwidth} m{0.37\textwidth}
		>{\centering\arraybackslash}m{0.035\textwidth} >{\centering\arraybackslash}m{0.035\textwidth}  }
	\toprule
	\textbf{Key Contribution} & \textbf{Methods \& Techniques Used} & \textbf{Year} & \textbf{Ref.} \\ \hline 
	\midrule
	
	Predicts e-commerce users' shopping intentions using LSTM recurrent neural networks. & Prediction, LSTM Recurrent Neural Networks & 2021 & \cite{Diamantaras2021} \\ 
	Provides an analysis of usage patterns and prediction through web usage mining techniques. & Pattern identification, Prediction, Clustering, Classification & 2024 & \cite{Dubey2024} \\ 
	Extracts patterns from proxy logs and predicts website requests. & Prediction, Fuzzy data mining, Fuzzy frequent mining & 2023 & \cite{Gangadwala2023} \\ 
	Proposes a neural network model for web page prediction using adaptive deer hunting and chicken swarm optimization. & Prediction, Apriori Algorithm, Chicken Swarm Optimization, Neural Network & 2022 & \cite{Gangurde2022} \\ 
	Develops a model for anomaly and intrusion detection using multi-demeanor fusion techniques. & Intrusion detection, Anomaly detection, Stochastic Latent Semantic Analysis & 2021 & \cite{Gupta2021} \\ 
	Enhances next-page prediction performance using web graphs and session reconstruction techniques. & Session reconstruction, Prediction, Bayesian network, Complete Session Reconstruction & 2023 & \cite{Jors2023} \\ 
	Uses a context-aware cohesive Markov model and Apriori algorithm for web usage pattern discovery. & Prediction, Cohesive Markov Model, Apriori Algorithm & 2022 & \cite{Luckose2022} \\ 
	Predicts user navigation patterns on websites using web usage mining techniques. & Prediction, Maximum frequent pattern, Classification & 2021 & \cite{Om2021} \\ 
	Proposes a system to predict users' learning styles on e-learning platforms. & Prediction, Spectral Clustering, Quadratic Support Vector Machine (E-SVM) & 2024 & \cite{Prashanth2024} \\ 
	Analyzes user behavior on e-commerce platforms and develops recommendation systems based on the findings. & Prediction, Random Forest classification, Event listeners & 2023 & \cite{Rajapaksha2023} \\ 
	Utilizes web access logs for semantic clustering to improve web prefetching accuracy. & Web prefetching, Prediction, Semantic clustering, Thesaurus (WordNet), SPUDK & 2024 & \cite{Setia2024} \\ 
	Improves user session clustering and prediction using semantic-based web session clustering methods. & Session clustering, Prediction, K-Means, Hierarchical Agglomerative Clust., Semantic distance & 2022 & \cite{Sowmya2022} \\
	
	\bottomrule
\end{tabular}
\end{table*}
\endgroup

In recent years, predictive modeling and anomaly detection have emerged as two prominent application areas within WUM \cite{Marcin2023, Hardik2023}. The main stages of web usage mining consist of data preprocessing, pattern discovery, and pattern analysis \cite{Suguna2017}. The first stage, data preprocessing, is essential for making raw web log data analyzable, involving tasks such as data cleaning, user identification, and session identification \cite{Panwar2018}. However, inaccuracies and errors in web log data can negatively impact the accuracy of analyses. For instance, incorrect session merging or user identification errors can lead to misleading results during modeling processes \cite{Sisodia2018}. Therefore, careful execution of data cleaning and session management processes is critical to the success of WUM.

Predictive modeling is a strategic method used to forecast users' future behaviors. These models utilize large-scale data analysis and machine learning algorithms to predict users' browsing habits, interactions, or purchasing tendencies \cite{Choudhary2023}. Such predictions provide valuable contributions, especially in dynamic fields like e-commerce and online services \cite{Ashraf2020}. On the other hand, anomaly detection identifies activities that deviate from standard user behavior patterns, providing crucial feedback in terms of security and performance \cite{Ilieva2021}. This section will examine recent developments in predictive modeling and anomaly detection within WUM, exploring studies and new approaches to enhance these processes' effectiveness.

\subsection{ Data preprocessing and session identification techniques}

Data preprocessing is critical in web usage mining as it optimizes large datasets and creates a more suitable environment for successful prediction and anomaly detection. Correcting errors in raw data and filtering out irrelevant information improves the accuracy of results significantly obtained from WUM processes. A recent study highlights the importance of data preprocessing in modeling user behavior \cite{Ali2020}. This study analyzed page views during user sessions, enabling more accurate session identification.

Session identification, an essential step in data preprocessing, plays a critical role in accurately analyzing users' navigation behavior. A new method for identifying web user sessions was developed, successfully generating all possible maximal paths \cite{Bayir2022}. This approach enabled more accurate structuring of user sessions, leading to superior results in subsequent page predictions. This process plays a significant role in making predictive modeling more efficient. Similarly, the Online Web Navigation Assistant (OWNA) analyzes real-time data streams during session identification, providing recommendations to users \cite{Ali2021}. This model optimizes user navigation behavior throughout sessions, improving the prediction of their actions on the web.

Tools used for data preprocessing in web usage mining also enhance the efficiency of processes. A hybrid approach has been developed that combines techniques like Ant Colony Optimization and Genetic Algorithm to improve classification accuracy during the data preprocessing stage \cite{Malik2021b}. Such tools filter errors and anomalies in large datasets, contributing to more successful prediction and anomaly detection outcomes.

\subsection{ Predictive modeling techniques}

Predictive modeling is one of the critical techniques used to anticipate users' next steps and predict potential actions on the web. One study demonstrated that the Compact Prediction Tree algorithm offers higher accuracy in predicting web pages than traditional methods such as k-nearest neighbor (k-NN) and decision trees \cite{Mani2020}. Similarly, recent research has highlighted the effectiveness of hybrid machine learning methods like Random Forest and Gradient Boosting in predicting web page transitions by utilizing both static and dynamic page features \cite{Dang2023}. Algorithms like these, employed to predict users' following pages, stand out as some of the most potent approaches in Web Usage Mining (WUM). Another study developed a web session reconstruction algorithm using a dynamic link repository \cite{Jors2023}. In this approach, web sessions were modeled graphically, and Bayesian networks were used to predict the next page, providing a dynamic prediction mechanism to optimize user movements across the web.

Another practical approach for predicting user behavior is the use of fuzzy logic-based algorithms. Using fuzzy data mining, one study analyzed proxy server log files to predict users' subsequent web requests \cite{Gangadwala2023}. By analyzing users' browsing frequency and behaviors, fuzzy association rules were created, enabling the accurate prediction of their next steps. Similarly, picture fuzzy logic was applied in a multi-criteria decision-making framework to evaluate website performance \cite{Karahan2024}. In another approach, fuzzy association rules were extracted from web data using learning automata, where trapezoidal membership functions (TMF) were used to optimize the time users spent on web pages, resulting in improved prediction accuracy \cite{Anari2021}. When classical machine learning methods fall short, these approaches offer a more flexible and adaptive prediction mechanism.

Clustering and classification techniques are also widely used in the predictive modeling process. One study employed hybrid methods combining classification techniques such as Random Forest and genetic algorithms to improve prediction accuracy \cite{Malik2021a}. These hybrid approaches, used in the context of WUM, allow for a more accurate classification of web log data. Similarly, another study used the fuzzy C-means algorithm to cluster user behaviors, making predictions based on these clusters \cite{Serin2022}. Following this line of research, an approach applied K-means and hierarchical clustering algorithms to group web sessions, extracting meaningful patterns from session data and predicting future user actions \cite{Sowmya2022}. Clustering algorithms are particularly effective in grouping user behaviors in large datasets and predicting future trends based on these groups.

\subsection{Anomaly detection approaches}

One key aspect of Web Usage Mining is the detection of anomalous user behaviors. Such anomalies can stem from various sources, including security threats, unusual user activities, or incorrect data inputs. Techniques that combine WUM with anomaly detection not only optimize user behavior analysis but also contribute to enhancing security measures. A modified hybrid method, combining PSO, GA, and K-Means, was developed for anomaly and misuse detection in computer networks \cite{Yuan2022}. This model detects abnormal behaviors in network traffic, allowing for minimizing security vulnerabilities.

Big data analytics plays a crucial role in anomaly detection within web mining. The IRPDP\_HT2 algorithm, developed as a scalable data preprocessing method based on Hadoop MapReduce, enables faster and more efficient detection of anomalies in large datasets \cite{Zhang2023}. Analyzing large-scale data sets for critical tasks such as robot detection proves to be an effective method for identifying abnormal activities. Similarly, dimensionality reduction techniques have been employed to detect anomalies in large datasets, providing scalable solutions for managing large volumes of web data \cite{Liu2022}. These approaches provide scalable solutions for managing large volumes of web data and detecting anomalies.

Hybrid methods are another approach to anomaly detection. A hybrid method combining the Grey Wolf Algorithm and CNN was developed to detect anomalous behavior in network data streams \cite{Wang2023}. Hybrid methods offer more flexible and efficient solutions for anomaly detection by integrating machine learning techniques with traditional approaches such as data compression \cite{Prasanth2015}.

\section{Methodology}\label{sec:3}

The CAWAL model \cite{Canay2024} was developed based on traditional application logging practices but expands this approach by integrating web analytics features. While conventional web analytics tools focus on capturing user interactions, modern tools have shifted toward more comprehensive data collection \cite{Canay2023}. The CAWAL model differentiates itself from other analytical tools by incorporating application logs and enhancing them with enriched log data and detailed user interaction analyses. This model was implemented on Sakarya University's institutional web application, the "Campus Automation Web Information System" (CAWIS) \cite{Canay2011}, where long-term access data was collected. The CAWIS system is architected as a web portal, utilizing separate subdomains for each service, which allows for tracking user interactions across different services and enhances the coverage and accuracy of the data gathered by the CAWAL model.

Compliance with Sakarya University's Internet Services Usage Policy Agreement was ensured during the data collection process. All necessary permissions were obtained, and anonymization methods were applied to protect user privacy. Timestamp data was altered to anonymize users' activities over time further. This adjustment is not expected to negatively affect the analysis results, as the study focuses on trends and behavioral patterns during specific periods rather than absolute timestamps. Every research stage was conducted meticulously to maintain participant privacy and ensure data security.

\subsection{ Integration of the CAWAL framework}

The CAWAL framework, designed to integrate with the web portal, detects user information and in-app events that third-party tracking tools fail to capture, storing the data in a structured format within a relational database \cite{Canay2024}. A data collection API continuously monitors exceptions, user flows, and state changes while also enabling the inclusion of application-specific data, such as form field entries in the tracking logs \cite{Canay2023}. Complete session tracking is maintained through persistent monitoring of the application servers. The CAWAL data collection API is initiated at the start of the web portal's code execution and integrates seamlessly to activate automatically with each page request. By integrating the API with the portal, the complexity of logging is abstracted from software processes, allowing developers to focus on core functions without being burdened by log management.

While the back-end code of the portal runs, the CAWAL data collection API operates within a multi-layered framework, gathering data in the background with each request. At the end of the portal's general interface template code, details such as page load times, database query delays, and error and warning messages are updated in the page view table via the API. This code-level tracking capability provided by CAWAL offers insights into applications, servers, and connections that would be otherwise unobtainable through traditional methods. This systematic approach enables the collection of comprehensive and unique data, helping to keep applications and systems under consistent monitoring.

The deployment of the framework in a real-world corporate web portal, encompassing a web farm with ten web servers and various web services spread across multiple subdomains, provides a distinctive approach to managing and analyzing extensive web traffic. CAWAL works harmoniously with a load-balancing mechanism configured to track operations across different servers simultaneously, ensuring system performance is maintained even during peak user activity periods. Using a shared NAS server to direct the session and configuration folders of the servers in the web farm provides an extra layer of consistency to the CAWAL deployment. This centralized storage solution guarantees continuity across the web farm, offering uniform and structured session data management. The architecture's support from load balancing and NAS servers enhances scalability, uninterrupted service delivery, and flexible solutions for complex web applications, thus improving overall system efficiency and its ability to adapt to the demands of the applications \cite{Canay2024}.

\subsection{ Data flow of the model}

\begin{figure*}[htb]
	\centering
	\includegraphics*[scale=0.60]{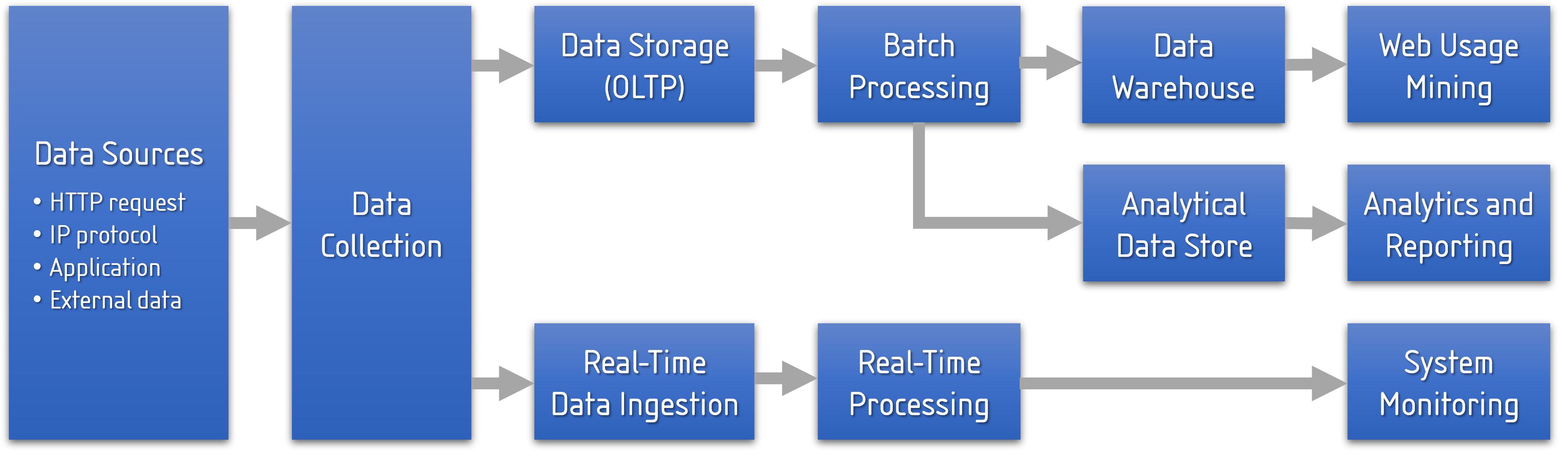}
	\caption{Data flow and processing in the CAWAL model.}
	\label{fig1}
\end{figure*}

Operational data generated during routine transactions in web applications is stored in write-intensive OLTP databases, which handle continuous data input and output operations \cite{Canay2023}. CAWAL implements a streamlined data model optimized for efficient analytics while minimizing storage overhead. \cref{fig1} presents a schematic detailing the data flow and processing steps in the CAWAL model.

The "Data Sources" section of the schematic illustrating the data flow and processing in the CAWAL model highlights various data sources, such as HTTP requests and network protocols. These data sources provide crucial information for monitoring user interactions on the portal in detail. For example, HTTP requests reveal which pages users visit and how long they stay on them, while network protocols provide system-level metrics such as server performance and the processing time for user requests. During the data collection phase, user and usage data is fused and prepared for processing, then stored in data storage systems. Temporal data gathered through the CAWAL framework is stored in a relational database in a homogeneous structure. These data can be used in real-time for system monitoring, as well as for anomaly detection and various predictions through machine learning algorithms.

The data stored in OLTP databases is processed during the batch processing stage for analytical insights and is later transferred to the data warehouse through ETL processes \cite{Canay2024}. The data housed in the warehouse then serves as a critical resource for future WUM analyses and predictive models. This approach minimizes the preprocessing required for WUM, ensuring the data is clean, consistent, and immediately ready for pattern discovery. The data collected through the CAWAL framework facilitates swift and accurate results in web usage mining processes, thanks to its high data accuracy and reliability.

\subsection{ Data preparation and enrichment}

Traditional web server logs typically provide limited information, focusing primarily on page visits and clickstreams. However, more detailed and specific data, such as time spent on a page or session login status, can only be captured at the application level or through enriched data sources. The CAWAL framework addresses this limitation by integrating web analytics with application logs to generate broader, more comprehensive datasets that are difficult to achieve with conventional methods. These comprehensive session and page view datasets, which include critical information such as the services accessed, the paths followed by users, and the time spent on each page, enable a more detailed tracking of user interactions within the web portal.

In addition to these fundamental details, enriched session data also captures other interactions performed by the user throughout the session. This data extends beyond just session start and end times, reflecting user behavior during the session. Information such as the number of pages visited, average time spent on pages, details of services used during the session, and page load times expand the scope of session data, allowing for deeper analysis. These insights form an essential foundation for understanding how and how often users engage with the portal's services and conducting thorough user journey analyses.

Enriched page view data includes not only basic page view information but also additional user attributes and session data. Knowing which session each page view belongs to, which services users accessed during that session, and how long they spent using these services enhances the depth of user experience analyses. Additional details such as browser type, IP location, and user type make performance evaluations more specific to diverse user segments. \cref{tab2} presents the fields and sample data from the enriched page view datasets to facilitate better understanding.

\begin{table*}[htb]
	\setlength{\parindent}{0pt} % Tablo girintisini sıfırla
	\centering
	\caption{CSV file format containing enriched pageview data.}
	\label{tab2}
	
	\begin{tabular}{p{0.2\textwidth} p{0.32\textwidth} >{\raggedright\arraybackslash}p{0.15\textwidth}} 
		
		\toprule
		\textbf{Field Name} & \textbf{Description} & \textbf{Sample Data} \\
		\midrule
		
		Detail\_ID & Pageview detail ID. & 89010871 \\
		Session\_ID & User's session ID. & 83665107 \\
		Detail\_Date\_Time & Request timestamp & 11.20.2022 13:01 \\
		User\_ID & User ID. & 184922 \\
		Current\_Login\_Status & Login status at the time of request. & 1 \\
		Session\_Login\_Status & Login status at the session. & 1 \\
		User\_Type & Type of portal user. & 6 \\
		Sex & User's sex. & 2 \\
		Age & User's age. & 18 \\
		Age\_Group & User's age group. & 1 \\
		User\_Language\_TR & User's browser language. & 1 \\
		User\_Location & User's IP location. & 1 \\
		Browser\_Type & User's browser type. & 1 \\
		Referer\_Type & Referrer type of the request. & 6 \\
		Server\_ID & Requested server ID. & 4 \\
		Service\_ID & Requested service ID. & 3 \\
		Page\_Duration & Page dwell time (sec). & 41 \\
		Page\_Load\_Time & Page load (generation) time (sec). & 0.122 \\
		\bottomrule
	\end{tabular}
\end{table*}

Enriched session and page view datasets encompass numerical and categorical fields and provide significant advantages for researchers and developers seeking to deeply analyze user behavior and portal performance. The data serves as a rich resource for critical analyses, such as understanding how users interact with services during sessions, how they transition between services, and how these transitions impact user engagement. This comprehensive and enriched analytical data can be leveraged to maximize the effectiveness of WUM activities, improve the accuracy of analyses, and yield more precise results.

Each field in the dataset facilitates detailed monitoring and analysis of user interactions. The session ID and user ID enable tracking user movements across the portal and evaluating behavioral changes over time. Metrics such as page load time and time spent on a page directly influence the performance of the web portal and provide critical insights for improving the user experience. These metrics also play a crucial role in analyzing behavioral differences among specific user groups. Demographic data such as user type, gender, and age group allow for the segmentation of user behavior, enabling more targeted and customized analyses.

\subsection{ Generation of data files}

The data collected by the CAWAL framework was processed through data selection and enrichment procedures, making it ready for the Web Usage Mining process. Using complex SQL queries, enriched session and page view data were created in data warehouses containing 8.5 GB of data spanning one month. This data was organized through Python scripts and stored in a view structure within the database. Subsequently, the numerical and categorical data were restructured for WUM applications. In the final stage, data for the time intervals specified in \cref{tab3} were extracted from these views and saved as comma-separated CSV files to facilitate easy processing by Python applications.

\begin{table*}[t]
	\setlength{\parindent}{0pt} % Tablo girintisini sıfırla
	\centering
	\caption{Enriched session and pageview data stored in CSV files and their properties.}
	\label{tab3}
	
	\begin{tabular}{p{0.65in}p{1.55in} >{\raggedright\arraybackslash}p{0.75in}>{\raggedleft\arraybackslash}p{0.6in}>{\raggedleft\arraybackslash}p{0.65in}}
		
		\toprule
		\textbf{Time Frame} & \textbf{Time Range} & \textbf{File Name .CSV} & \textbf{Record Count} & \textbf{File Size (MB)} \\
		\midrule
		
		1-day & 2022-11-22 00.00 - 23.59 & va\_page2 & 787,637 & 66.3 \\
		1-week & 2022-11-21 - 2022-11-27 & va\_sess4 & 514,879 & 99.2 \\
		1-month & 2022-11-01 - 2022-11-30 & va\_sess5 & 1,220,916 & 235.0 \\		
		
		\bottomrule
	\end{tabular}
\end{table*}

These structured datasets are critical for analyzing users' interactions with the portal's various services. For instance, the daily page view dataset contains 787,637 records, while the monthly session dataset consists of 1,220,916 records. Enriching these datasets is particularly essential for detailed analysis of user behavior. This process enables the identification of varying needs across different user segments, allowing for service improvements to be optimized accordingly. Metrics such as page load time provide insights into performance indicators directly affecting user engagement. These large datasets, which are extremely valuable for analyzing user behaviors over specific periods and examining usage patterns during special events, can be used to identify performance bottlenecks and develop optimization strategies. Storing the data in CSV format enables researchers to process it quickly and efficiently, providing high accessibility for web usage mining and other analytical applications.

\subsection{ Feature engineering and model training}

\begin{figure*}[hb]
	\centering
	\includegraphics*[scale=0.60]{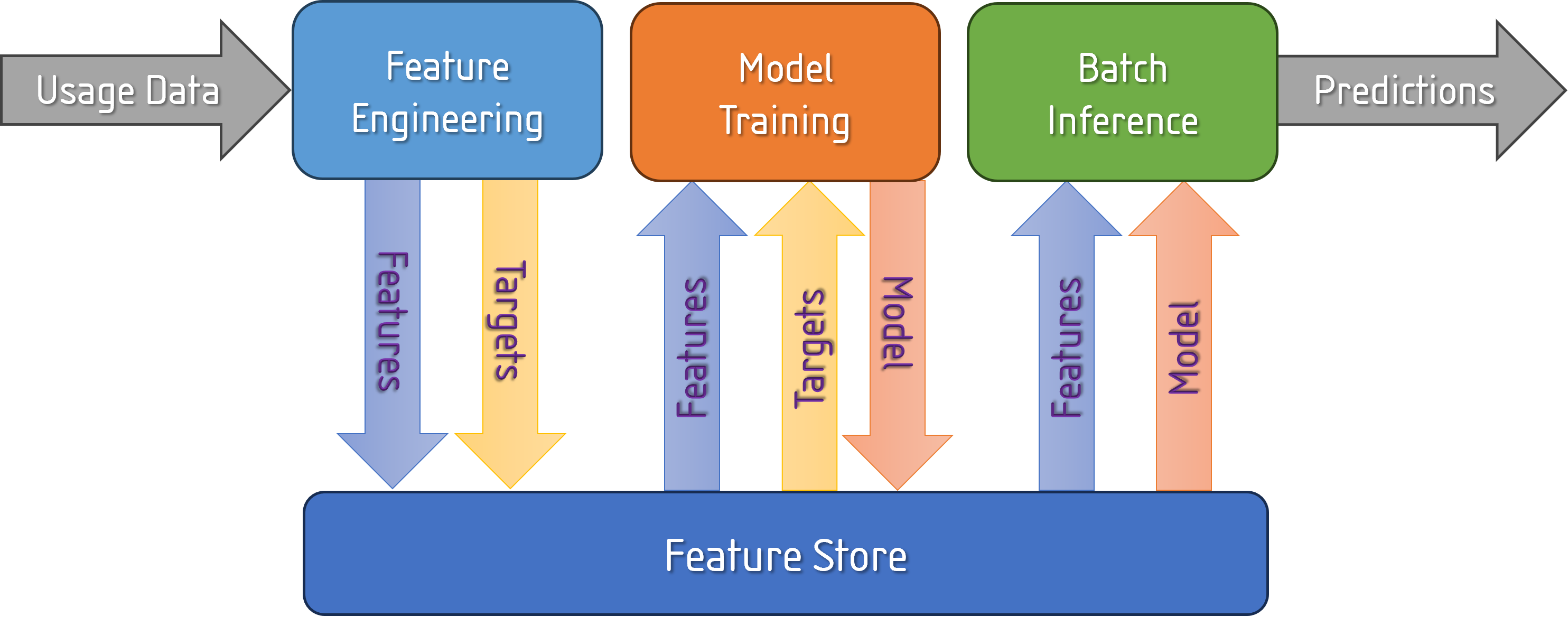}
	\caption{Transformation of web usage data into predictions.}
	\label{fig2}
\end{figure*}

The success of data mining processes relies heavily on effectively processing raw data and transforming it into meaningful features. This process involved comprehensive tasks such as cleaning session data, selecting relevant features, and preparing these features for modeling. \cref{fig2} conceptually illustrates how usage data is transformed through feature engineering into prediction models.

During the feature engineering phase, attributes that accurately reflect the complexity of user interactions were identified, and these features were used to train prediction models to reveal patterns and trends in the data. The specific features selected for the models are detailed in the analysis section of the study. In the modeling phase, a Random Forest Classifier was employed to predict users' likelihood of abandoning the system, while a Gradient Boosting Classifier was used to predict the last service accessed before abandonment. These models were trained and tested on extensive session and pageview datasets.

In order to predict the probability of a user accessing a specific service, four different models were applied: Gradient Boosting, Random Forest, Support Vector Machine, and Logistic Regression. The performance of these models was enhanced through hyperparameter optimization, ensuring the best parameters were selected. Additionally, for anomaly detection based on server and page load times, the Isolation Forest algorithm was used to identify potential anomalous user behaviors and performance deviations. The feature pool was structured to include extensive attributes, such as enriched session and page data, user demographic information, browser and device types, and system performance metrics. This centralized repository played a crucial role in the process of cleaning, selecting, and preparing data for analysis, ensuring consistent and reliable results during the modeling phase.

\section{Predictive analysis and findings}\label{sec:4}

Four analyses were conducted within the scope of this research using enriched session and page view data obtained through the CAWAL framework, including three aimed at prediction and one at anomaly detection. The first three analyses, aimed at enhancing the efficiency of the web portal, focus on predicting users' exit methods, the last services they abandon, and their access to a specific portal service. On the other hand, the analysis for anomaly detection focuses on identifying abnormalities based on page load times on web farm servers, providing essential insights into the security and performance of the portal. The findings obtained from these analyses provide a strong foundation for strategies to enhance the portal's effectiveness by enabling a better understanding of user behavior.

\begin{table*}[htb]
	\setlength{\parindent}{0pt} % Tablo girintisini sıfırla
	\centering
	\caption{Class-wise and overall performance metrics of the multiple classification model.}
	\label{tab4}
	
	\begin{tabular}{p{1.6in}>{\raggedleft\arraybackslash}p{0.65in} >{\raggedleft\arraybackslash}p{0.65in}>{\raggedleft\arraybackslash}p{0.65in}>{\raggedleft\arraybackslash}p{0.65in}}
		
		\toprule
		\textbf{Class} & \textbf{Precision} & \textbf{Recall} & \textbf{F1 Score} & \textbf{Support} \\
		\midrule
		
		0 (direct leave) & 0.96 & 0.91 & 0.93 & 163,187 \\
		1 (logout button) & 0.92 & 0.96 & 0.94 & 192,866 \\
		2 (notification window) & 0.77 & 0.86 & 0.81 & 10,222 \\
		\midrule
		Weighted Avg. & 0.93 & 0.93 & 0.93 & 366,275 \\
		
		\bottomrule
	\end{tabular}
\end{table*}

\subsection{Prediction of methods for leaving the system}

This analysis, which focuses on predicting how users will leave the system, is hugely significant in understanding interactions and behavioral patterns. The prediction model used in the study was trained using a Random Forest Classifier \cite{Manzali2023} with enriched session data. The training utilized several user and session-related attributes, including User\_Type, Sex, Age, User\_Language\_TR, User\_Location, Browser\_Type, Landing\_Srv\_ID, Exit\_Srv\_ID, Session\_Login\_Status, Page\_Count, Service\_Count, Total\_Session\_Duration, Avg\_Page\_Duration, Total\_Page\_Load, p\_gate, p\_mail, p\_obis, p\_abis, p\_pbis, and p\_menu.

A one-month dataset of 1,220,916 enriched session data rows was split into 70\% for training and 30\% for testing, yielding highly successful prediction results. \cref{tab4} displays the precision, recall, and F1 score values for each class, as well as the support counts. The weighted average represents the model's overall performance, calculated based on all classes' proportions.

\begin{figure*}[b]
	\centering
	\includegraphics*[scale=0.54]{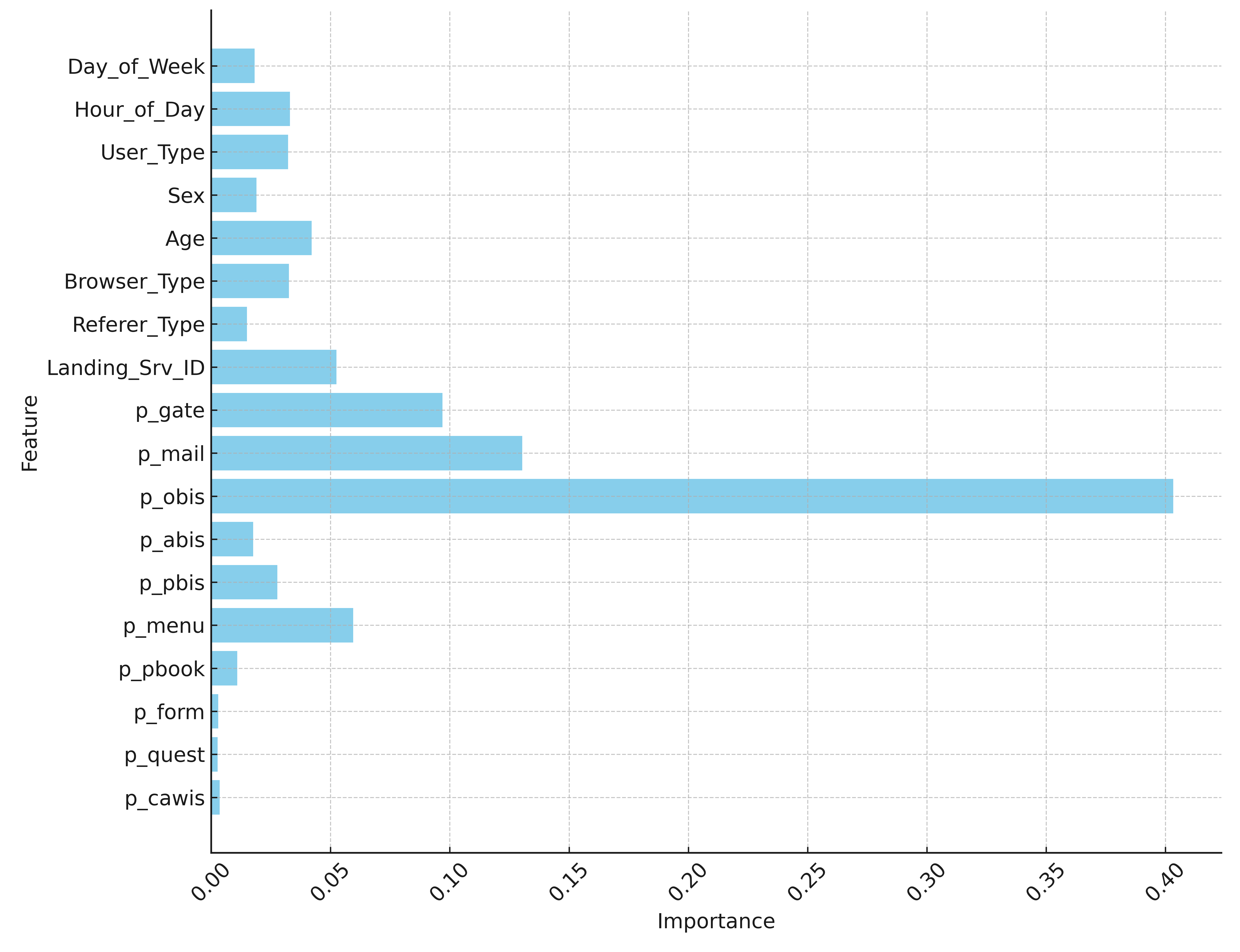}
	\caption{Feature importance scores in the predictive model.}
	\label{fig3}
\end{figure*}

When the model's performance is examined to identify individual classes and overall predictions, it demonstrates remarkable accuracy. This finding reflects the model's ability to distinguish specific classes and its overall predictive success across the dataset. The high precision and good recall values obtained for direct departures (Class 0) indicate the model's effectiveness in identifying this class. Similarly, for secure exits via the exit button (Class 1), the high precision and even higher recall values demonstrate the model's strong performance in predicting this class.

The F1 scores for both classes indicate that the model performs a balanced performance in predicting these classes. On the other hand, the precision and recall values for Class 2 (timeout notification window) are lower than other classes. However, the F1 score obtained for this class suggests the model's performance in predicting this class is acceptable. It can be inferred that the relatively minor number of examples for this class leads to slightly lower performance results compared to the other classes. 

The evaluation of the model's performance, mainly through the weighted average F1 score, demonstrates that it achieves balanced and high accuracy across all classes. This effectiveness is further supported by the enriched data provided by CAWAL, which enhances the model's predictive capabilities for different exit methods. These findings highlight the model's robustness in handling diverse user behaviors and its potential application for improving system design and user experience.

\begin{table*}[b]
	\setlength{\parindent}{0pt} % Tablo girintisini sıfırla
	\centering
	\caption{Performance comparison of various classification models for predicting mail service usage.}
	\label{tab5}
	
	\begin{tabular}{p{1.3in}>{\raggedright\arraybackslash}p{1.22in}>{\raggedright\arraybackslash}p{1.22in}>{\raggedright\arraybackslash}p{1.22in}p{0.75in}}
		\toprule
		\textbf{Metric / Model} & \textbf{Gradient} \newline \textbf{Boosting} & \textbf{Random} \newline \textbf{Forest} & \textbf{Support} \newline \textbf{Vector Mach.} & \textbf{Logistic} \newline \textbf{Regression} \\
		\midrule
		 Best Parameters & Learning Rate: 0.1, \newline Max Depth: 3, \newline N Estimators: 100 & Max Depth: 10, \newline Min Samp. Leaf: 2, \newline Min Samp. Split: 5, \newline N Estimators: 100 & C: 10, \newline Kernel: 'rbf' & C: 10 \\
		Average CV Score & 0.9182 & 0.8947 & 0.9184 & 0.9143 \\
		Accuracy & 0.9252 & 0.9240 & 0.9238 & 0.9178 \\
		Precision & 0.9234 & 0.9218 & 0.9214 & 0.9156 \\
		Recall & 0.9252 & 0.9240 & 0.9238 & 0.9178 \\
		F1 Score & 0.9188 & 0.9176 & 0.9174 & 0.9094 \\
		\bottomrule
	\end{tabular}
	
\end{table*}

\subsection{Prediction of the last abandoned service}

This part of the study focuses on predicting the service through which users will leave the system. For this purpose, the model is trained with a gradient boosting classifier \cite{Omari2023} using various user and session features from a comprehensive dataset containing one-week session information. In the model's training, 0.1 was chosen as the learning rate, three as the maximum depth, and 100 as the number of predictors. The features used in the model and their importance on the predictive ability of the model are presented in detail in \cref{fig3}.

The details of the features used in the model and their impact on the model's predictive ability highlight the depth and comprehensiveness of these analyses. Notably, the features p\_obis (0.4033), p\_mail (0.1304), and p\_gate (0.0969) stand out as the most influential factors that significantly enhance the model's prediction accuracy. These p\_ prefixed features represent the number of pages users visit within specific services. These findings indicate that users' interactions with the "obis, "mail, "and "gate " services play a crucial role in determining their exit points from the web portal. The high importance of these features dramatically increases the predictability of users' interactions with the system, providing valuable insights for optimizing the web portal's performance and enhancing the user experience.

According to the model's test results, the accuracy is measured at 0.9557, precision at 0.9561, recall at 0.9557, and the F1 score at 0.9555. These results demonstrate that the model is highly effective in predicting through which service users will exit the system. These findings prove that using the data provided by the CAWAL framework, it is possible to predict with high accuracy the last service through which users will leave the portal.

\subsection{Prediction of access to a specific service}

WUM methods are crucial for gaining deeper insights into portal interactions. Prediction models can be used to determine whether a user will access a particular portal service in a session. Various session information such as Log\_Date\_Time, Log\_Date, User\_Type, Sex, Age, Avg\_Page\_Duration, User\_Language\_TR, User\_Location, Browser\_Type, and Referer\_Type was trained with four different models to perform service access prediction. Hyperparameter optimization was performed to achieve the best performance, and the optimal parameters of the four models were determined. \cref{tab5} presents the performance of the classification models and their parametric configurations in a comparative manner. 

This representation reveals the models' differences by demonstrating the best parameter combinations and performance metrics, such as average cross-validation (CV) score, accuracy, precision, recall, and F1 score side-by-side. When the model performance metrics are analyzed, it is seen that all four models exhibit high accuracy rates and balanced F1 scores. The fact that the Random Forest and Gradient Boosting models stand out in terms of both accuracy and F1 score indicates that these two models have a better generalization capability on the dataset. The combination of values identified as the best parameters of the Random Forest model shows that the model manages its complexity and learning ability in a balanced way.

The Logistic Regression model \cite{He2023} achieved a high Average Cross-Validation Score with the best C parameter but had a slightly lower F1 score, indicating potential difficulty in discriminating between certain classes. In contrast, the Support Vector Machine (SVM) model \cite{Guo2023} demonstrated high generalization capability with its RBF kernel and C parameter. Both models offer a balanced approach to the classification problem. The Gradient Boosting model's performance improved by carefully tuning the learning rate, maximum depth, and number of predictors, successfully capturing the dataset's complex structures with high accuracy and F1 score. While all models show high accuracy, Gradient Boosting and Random Forest particularly excel in capturing complex patterns. Despite strong cross-validation performance, the Logistic Regression model may slightly struggle with class separability, suggesting that tree-based models might better fit scenarios where class distinction is critical.

The four models applied have produced effective results for this classification problem and have accurately predicted whether a user will access a service based on specific features. The fact that each model's parameter settings were adjusted to provide the best results suited to the system's data structure demonstrates the significant role of hyperparameter optimization. This optimization has enhanced each model's performance, leading to more precise and reliable predictions. Specifically, Gradient Boosting and Random Forest models have effectively utilized the selected session-specific features to capture the complex interactions in the dataset, achieving high accuracy and balanced F1 scores. These results indicate that the detailed information provided by CAWAL, including user demographics, behavioral patterns, and contextual factors, has significantly improved prediction performance. Beyond the primary interaction data provided by traditional web server logs, these enriched features have enabled the models to understand the complex aspects of user behavior better and make more accurate predictions.

\begin{figure*}[b]
	\centering
	\includegraphics*[scale=0.73]{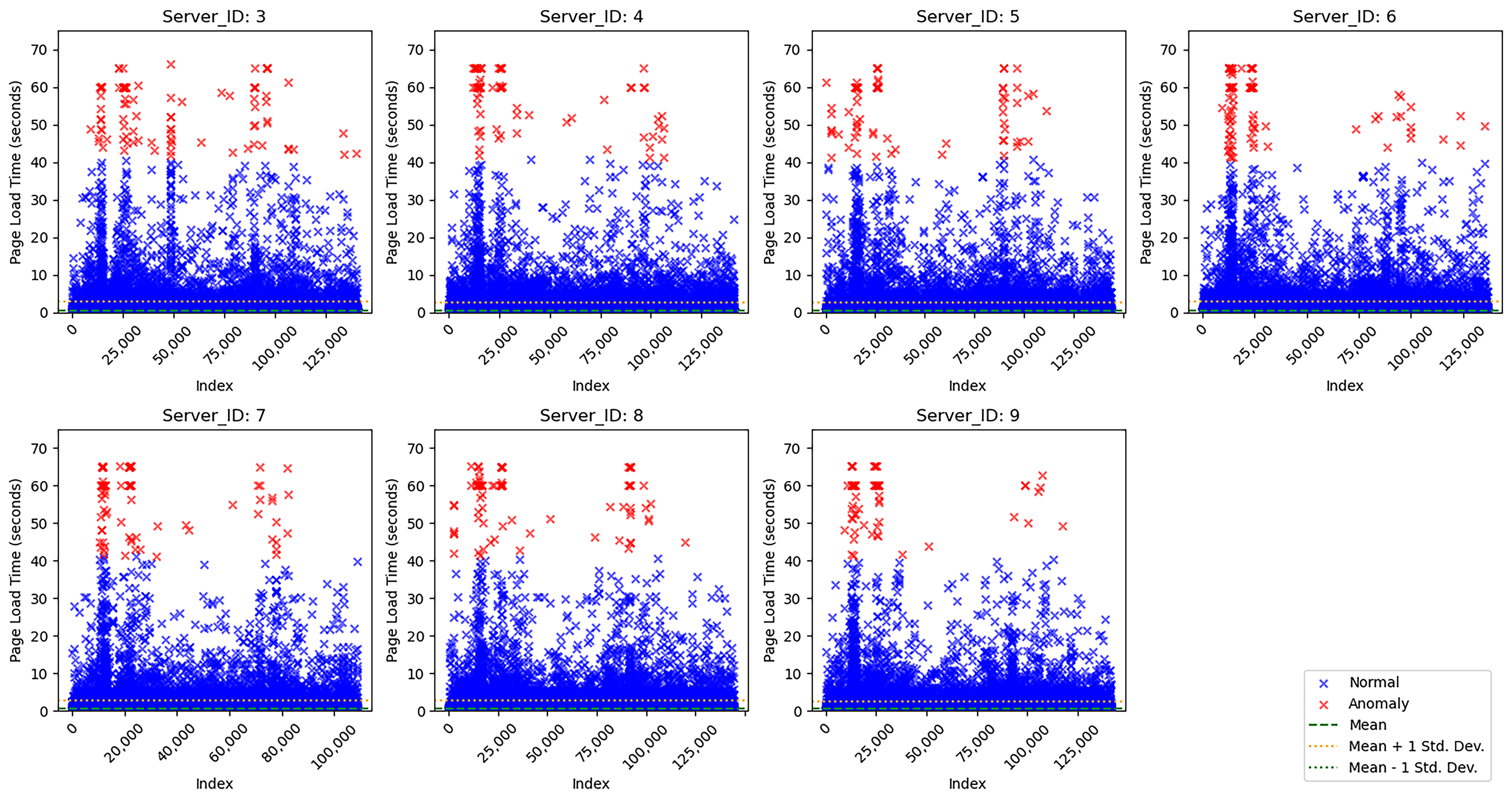}
	\caption{Visualization of anomalies based on page load times across seven servers in the web farm.}
	\label{fig4}
\end{figure*}

\subsection{Server and page load time-based anomaly detection}

In a web farm architecture, the systematic analysis of server performance and page load times is crucial for detecting anomalies within applications and servers. These issues can arise from software bugs, hardware malfunctions, network latency, interactions with databases or other connected systems, denial-of-service (DoS) attacks, or server configuration problems. In this analysis, anomaly detection was performed using the Isolation Forest (iForest) algorithm, an effective method for identifying outliers in a dataset based on isolation principles, as demonstrated in a recent study \cite{Alshehari2023}. The Isolation Forest works by recursively partitioning the data space through multiple binary trees, known as isolation trees or iTrees. The main idea behind this method is that anomalies are relatively sparse and distinct, which makes them easier to isolate. To achieve this, the mathematical techniques used in the anomaly detection process are explained in detail below.

One such technique is the Isolation Forest algorithm, which detects anomalies by isolating data points using random partitioning. The isolation process begins by randomly selecting a feature and then selecting a split value between the minimum and maximum values of that feature. This process recursively repeats, creating partitions that progressively isolate individual data points. The key metric in Isolation Forest is the \textit{path length} \( h(x) \), defined as the number of edges traversed from the root node to the terminating node for a specific data point \( x \) in an isolation tree. Anomalous data points tend to have shorter path lengths because they are easier to isolate compared to normal points. 

For the dataset with \( n \) instances, the average path length \( c(n) \) of unsuccessful searches in a Binary Search Tree can be approximated by:

\[c(n) = 2 \cdot H(n - 1) - \frac{2(n - 1)}{n}\]

where \( H(n - 1) \) is the harmonic number, estimated as:

\[H(n - 1) = \ln(n - 1) + \gamma\]

and \( \gamma \) is the Euler-Mascheroni constant (\( \gamma \approx 0.5772 \)).

The anomaly score for a data point \( x \) is calculated based on its average path length \( h(x) \) across all isolation trees in the forest. The score \( s(x, n) \) is given by:

\[s(x, n) = 2^{ - \frac{h(x)}{c(n)} }\]

Specifically, \( s(x, n) \) ranges from 0 to 1, where points with a score close to 1 are considered anomalies due to their shorter path lengths. Isolation Forest is advantageous because of its linear time complexity with a low constant and its ability to handle high-dimensional data efficiently. Moreover, it does not require prior knowledge of the anomaly ratio within the dataset, making it a versatile tool for various applications.

Considering these advantages, we applied Isolation Forest to analyze the 24-hour page view data and load times of requests distributed through load balancing in a multi-server architecture. This approach enabled us to detect anomalies in web traffic patterns, which could indicate issues such as server overloads, network delays, or potential security threats. The results of the analysis performed with the model trained using the Isolation Forest algorithm are presented in \cref{fig4}.

The visualization illustrates anomalies detected based on page load times across seven diverse servers. Each graph shows the distribution of page load times, which refers to the time the back-end code takes to generate the page over time, and highlights anomalies detected for the web server identified by its Server\_ID. The "Y" axis of the sub-graphs represents the page load times, while the "X" axis represents the "Index" value, corresponding to the number of page views observed over time on each server. In the graphs, regular observations are marked by blue 'x' symbols, while anomalies, where the load time is significantly longer than expected, are indicated by red markers.

Data analysis from a weekday with heavy system usage reveals several requests with high page load times. While the number of anomalies varies across the servers, the distribution appears generally balanced. Anomalies were detected when page load times deviated more than one standard deviation from the average. An exceptionally high number of anomalies observed on servers 3 and 6 suggests that these servers experienced more performance issues than others, especially in their interactions with related components such as other web servers, database servers, LDAP, and mail servers. These problems may have led to noticeable delays in page load times and even page timeout occurrences.

The delays in connections to heavily loaded servers and disruptions in queries performed on other application databases are among the primary causes of the anomalies. Performance issues in interactions with other components, especially on specific servers, cause significant delays and timeouts in page load times. The data provided by the CAWAL framework plays a critical role in detecting and analyzing these issues, contributing significantly to the overall evaluation of system performance.

\section{Discussion}\label{sec:5}

The CAWAL framework has successfully overcome the limitations of conventional Web Usage Mining methods by analyzing user behaviors in web portals. Traditional approaches rely solely on server logs, resulting in superficial and limited analysis of user interactions. Due to the limited data sources, these constraints make it difficult to understand user behavior. For instance, one study  \cite{Ali2020} attempted to model user behavior by preprocessing web data, but the diversity of the data remained restricted. Similarly, a new method for constructing user sessions was proposed \cite{Bayir2022}, but it also faced the limitations of data richness from weblogs.

The CAWAL framework, developed as a web analytics solution, combines application logs with web analytics to provide a richer and more comprehensive dataset for large-scale web portals. This innovative approach offers significant advantages in multidimensional data integration and richness compared to weblogs, widely used as a data source in the literature for WUM. The framework enhances data quality by addressing the limitations of relying solely on weblogs. Improved data quality and diversity enable detailed and precise analysis of user interactions, resulting in impressive outcomes in analyses, where predictive models achieved over 92\% accuracy and server-based anomaly detection yielded significant findings.

In this study, predicting the last service accessed before users exit the system was successfully achieved using the Gradient Boosting algorithm, with an accuracy of 95.61\% and an F1 score of 95.55\%. Detailed user and session data significantly enhanced the model's capacity to capture complex behavioral patterns. In a similar study, LSTM networks were employed to predict e-commerce users' shopping intentions \cite{Diamantaras2021}, but the generalization capacity was limited due to insufficient data integration. The comprehensive data integration offered by CAWAL addresses this gap in the literature by improving the accuracy of predictive models. These results enable strategic decisions to enhance critical services in the system by accurately predicting the points at which users exit the portal.

The prediction of users' exit methods was effectively achieved using the Random Forest model, which demonstrated a weighted average F1 score of 93\%, accurately forecasting different exit methods. The richness of the data provided by CAWAL enhanced the model's capacity for accurate predictions. A similar approach aimed to automatically extract users' browsing patterns \cite{Mani2020}, but the accuracy achieved did not reach the levels provided by CAWAL. These predictions can be considered a strategic tool for system design and user experience optimization. The obtained results demonstrate that CAWAL's depth and accuracy in analyzing user behavior contribute significantly to improving the performance of web portals.

The success of models predicting users' access to specific services was also remarkable, thanks to CAWAL's rich data sources. Comparing model results revealed that hyperparameter tuning was critical in accurately capturing the complex structures within the dataset. The Gradient Boosting model, with carefully tuned parameters such as learning rate, maximum depth, and the number of predictors, captured complex patterns with 91.88\% accuracy and a 91.76\% F1 score. The Random Forest model also performed well, with an accuracy of 92.40\% and a 91.76\% F1 score. Although the Logistic Regression model delivered effective results with a high cross-validation score, it performed relatively lower in the F1 score, indicating its limitations in distinguishing certain classes. The SVM model demonstrated balanced success with an RBF kernel and C parameter. These findings suggest that model selection and parameter tuning are crucial for obtaining results in classification problems, particularly with complex datasets. A study applying machine learning approaches to predict learning styles in e-learning platforms \cite{Prashanth2024} achieved lower accuracy due to limited data diversity.

The high predictive success achieved across these three analyses strongly supports the initial hypothesis. The enriched datasets provided by the CAWAL framework have improved the accuracy of machine learning-based predictive models in large-scale architectures. The models' high accuracy and F1 scores demonstrate how CAWAL's data integration and collection capabilities contribute to effectively modeling complex user behaviors. While previous studies used various methods to determine user access behaviors on the web \cite{Alhaidari2020, Gangadwala2023, Soewito2023}, these approaches were limited in terms of data diversity and richness because they mainly relied on weblogs. CAWAL provides deeper analysis capabilities than traditional WUM methods, overcoming these limitations and enabling higher accuracy rates. Analyzing details such as session ID and page load times enabled more accurate predictions of user behavior, resulting in significant improvements in the system. These results indicate that CAWAL is highly effective in performance optimization and enhancing user engagement in high-traffic web portals.

The successful outcomes of anomaly detection further confirm the second hypothesis. The CAWAL framework enhances anomaly detection in web farms and multi-server architectures, helping to detect operational disruptions early and maintain system stability. The anomalies detected during the analysis could be due to factors such as insufficient server resources, software errors, or high traffic volumes. For example, the delays observed in email and database servers are thought to be caused by resource bottlenecks or excessive demand. Such analyses are valuable for improving system efficiency and developing proactive solutions for future needs. While a previous study examined anomaly detection in networks using WUM techniques \cite{Gupta2021}, CAWAL's success in large-scale, multi-server environments provides a broader scope. The enriched datasets provided by CAWAL accelerate anomaly detection, allowing potential issues to be identified earlier.

However, the CAWAL framework does have some limitations. One limitation is that the datasets used in the study focus on a specific period and user group, which may limit the generalizability of the findings. The framework's effectiveness has not been thoroughly tested in environments demonstrating diverse user behaviors, such as various industries, e-commerce, and mobile platforms. This situation highlights the need for further investigation into how CAWAL handles broad data diversity and models various user behaviors.

Additionally, although the CAWAL framework can process large datasets, aspects of the data processing pipeline require optimization in terms of time and computational costs. Specifically, multi-server systems' data collection and analysis processes demand substantial computational power and time, which may pose challenges for real-time applications. Furthermore, the performance of the machine learning models heavily depends on the scope and quality of the datasets. The effectiveness of these models may decrease with limited or imbalanced datasets.

Moreover, in environments where comprehensive and detailed user information is tracked, such as with CAWAL, stricter measures should be taken to ensure the security and privacy of the data, considering the associated risks. Handling such sensitive data requires the adoption of robust encryption methods and compliance with data protection regulations, such as GDPR \citep{Negriribalta2024}, to prevent data breaches and unauthorized access, particularly in multi-server systems where vulnerabilities may increase.

\section{Conclusion}\label{sec:6}

The CAWAL framework has demonstrated superior performance in predicting user behaviors and detecting anomalies in web portals, surpassing traditional methods. Integrating data collected from four diverse sources with web analytics and application logs has enabled more precise and in-depth analysis of user interactions, particularly in large-scale architectures with multi-server and multi-subdomain structures. This approach provides a reliable solution for organizations that find web server logs insufficient or are unwilling to share user access data, particularly in web farms and multi-server environments. The model's secure and structured data collection mechanism has proven to be an efficient resource for web usage mining and machine learning models, even in high-traffic, large-scale systems. The findings demonstrate that the richness and integration capabilities provided by the framework deliver high accuracy and efficiency in both predictive models and anomaly detection processes.

The analysis results show that the data stream provided by the CAWAL framework accelerates the preprocessing stage and strengthens predictive models. In large-scale architectures, more accurate forecasting of user behaviors and system performance optimization confirm the model's successful application in practice. Furthermore, the successful results obtained from anomaly detection analyses reveal that CAWAL optimizes anomaly detection in web farms and multi-server architectures, thereby improving system efficiency and security. The swift and effective handling of detected anomalies reduces operational risks and enhances overall system performance. These findings support both hypotheses, confirming that the CAWAL framework is not just a theoretical innovation but a tangible and applicable framework for performance optimization and security enhancements in large-scale and complex web portals.

Future research should focus on extensive testing of the model across different industries and larger datasets. These tests, aimed at increasing CAWAL's generalizability and applicability to diverse user groups, will further strengthen its flexibility and versatility, demonstrating its effectiveness across various data environments and architectural structures. Particularly in the e-commerce, finance, and healthcare sectors, the concrete effects of the framework's data integration, model accuracy, and system optimization in large-scale, multi-server systems should be evaluated. Such studies will validate CAWAL's broad application potential and contribute to identifying new approaches to predictive modeling and anomaly detection within the scope of web usage mining.

\section*{CRediT authorship contribution statement}\label{CRedit}
\textbf{Özkan Canay:} Conceptualization, Methodology, Investigation, Software, Data curation, Validation, Visualization, Writing – original draft, Writing – review \& editing. \textbf{Ümit Kocabıçak:} Conceptualization, Supervision.

\section*{Declaration of generative AI and AI-assisted technologies in the writing process}\label{decAI}
During the preparation of this work the authors used DeepL, ChatGPT, and Grammarly in order to English translation and editing. After using these tools/services, the authors reviewed and edited the content as needed and take full responsibility for the content of the publication.

\section*{Ethics Statement}\label{ethics}
The data utilized in this research were collected from the CAWIS web portal following legal statutes and Sakarya University's regulations. Necessary permissions were secured from the institution, and various data anonymization techniques were applied throughout the study to ensure user privacy and data security.

\section*{Consent Statement}\label{consent}
Consent for data usage was secured through the Internet Services Usage Policy Agreement, which all portal users approved.

\section*{Declaration of competing interest}\label{declare}
The authors declare that they have no known competing financial interests or personal relationships that could have appeared to influence the work reported in this paper.

\section*{Data availability statement}\label{data_availability}
The data cannot be made available for public access due to confidentiality agreement restrictions.

\bibliographystyle{elsarticle-num}
\bibliography{refs}

\end{document}